\documentclass[11pt,a4paper]{article}
\usepackage{times,latexsym}
\usepackage{url}
\usepackage[T1]{fontenc}

\usepackage[acceptedWithA]{tacl2021v1}
\usepackage{tacl2021v1}

\usepackage{xspace,mfirstuc,tabulary}

\newif\iftaclinstructions
\taclinstructionsfalse 
\iftaclinstructions
\renewcommand{\confidential}{}
\renewcommand{\anonsubtext}{(No author info supplied here, for consistency with
TACL-submission anonymization requirements)}
\newcommand{\instr}
\fi

\iftaclpubformat 

\else

\fi

\usepackage{graphicx}

\title{Start Making Sense(s): A Developmental Probe of Attention Specialization Using Lexical Ambiguity}

\author{
  Pamela D. Rivi\`ere
  \\
  University of California, San Diego
  \\
  \texttt{pdrivier@ucsd.edu}
  \And
  Sean Trott 
  \\
  Rutgers University - Newark
  \\
  \texttt{sean.trott@rutgers.edu}
}

\date{}

\begin{document}
\maketitle
\begin{abstract}
 Despite an in-principle understanding of self-attention matrix operations in Transformer language models (LMs), it remains unclear precisely how these operations map onto \textit{interpretable} computations or functions---and how or when individual attention heads develop specialized attention patterns. Here, we present a pipeline to systematically probe attention mechanisms, and we illustrate its value by leveraging lexical ambiguity---where a single word has multiple meanings---to isolate attention mechanisms that contribute to word sense disambiguation. We take a ``developmental'' approach: first, using publicly available Pythia LM checkpoints, we \textit{identify} inflection points in disambiguation performance for each LM in the suite; in $14M$ and $410M$, we identify heads whose attention to disambiguating words covaries with overall disambiguation performance across development. We then \textit{stress-test} the robustness of these heads to stimulus perturbations: in $14M$, we find limited robustness, but in $410M$, we identify multiple heads with surprisingly generalizable behavior. Then, in a \textit{causal analysis}, we find that ablating the target heads demonstrably impairs disambiguation performance, particularly in $14M$. We additionally reproduce developmental analyses of $14M$ across all of its random seeds. Together, these results suggest: that disambiguation benefits from a constellation of mechanisms, some of which (especially in $14M$) are highly sensitive to the position and part-of-speech of the disambiguating cue; and that larger models ($410M$) may contain heads with more \textit{robust} disambiguation behavior. They also join a growing body of work that highlights the value of adopting a developmental perspective when probing LM mechanisms.


\end{abstract}


\section{Introduction}

Transformer-based language models (LMs) \citep{vaswani2017attention} have proven adept both at the primary language modeling objective and at a variety of downstream tasks. A core innovation of Transformers is the attention block, which consists of a series of parallel ``attention heads'': sets of Query, Key, and Value matrices that each transform token vectors (or ``embeddings'') according to linguistic context. Yet despite the success of these systems, there remains limited mechanistic understanding of how specific model components give rise to observable behaviors or implement particular functions; developing such understanding is the central goal of research on model interpretability \citep{sharkey2025open, pmlr-v162-geiger22a, mueller2025questrightmediatorsurveying}. In particular, a growing body of work suggests that different attention heads learn to ``specialize'' (ie., over the course of pre-training) in which kinds of context they attend to \citep{olsson2022context,wanginterpretability,merullo2024talking, park2025does}.  

Here, we investigate contextualization mechanisms in the Transformer LM architecture by combining careful psycholinguistic \textit{experimental design} and targeted \textit{editing of LM weights} at multiple pre-training LM checkpoints. The former allows us to identify key ``developmental milestones'' associated with improvements in contextualization over the course of pre-training, and also isolate candidate attention heads that integrate information from disambiguating cues. The latter allows us to assess the causal influence of these components in disambiguation performance.

Specifically, we leverage a dataset of human relatedness judgments of (English) ambiguous words \citep{trott2021raw} to probe which components of Pythia-$14M$ and Pythia-$410M$ \citep{biderman2023pythia} contribute to contextualization---and when these mechanisms develop throughout pretraining. As noted below (Section \ref{sec:related}), ambiguity is a useful probe for several reasons: it is exceedingly common \citep{rodd2004modelling}, word sense disambiguation is an established task of interest in the field of natural language processing (NLP) \citep{haber2020word}, and the process of disambiguation is a specific sub-case of the more general challenge of \textit{contextualization}---one that importantly allows us to vary the surrounding context while keeping the target token identity the same. This work extends the study of self-attention and linguistic contextualization, and underscores the value of deploying carefully designed variations in stimuli---coupled with causal manipulations of isolated attention heads---to sketch any given head's \textit{functional scope} and assess its contributions to model behavior.\footnote{All code and data required to reproduce the analyses for each model at each checkpoint is available at the following Github repository: \url{https://github.com/seantrott/entangled_meanings}}

\section{Related Work}\label{sec:related}

Prior work has successfully adopted a ``developmental approach'' to investigate the emergence and nature of Transformer attention head specialization \citep{chensudden}. Focusing on the evolution of self-attention over the course of pre-training, \citet{olsson2022context} identified ``induction heads'': attention heads that selectively track repetitions of a target token in the preceding context---as well as the tokens that follow---in order to produce correct next-token completions. The emergence of inductive attentional patterns coincides with marked improvements to in-context learning \citep{olsson2022context}, with subsequent work establishing a causal role for induction heads in successful completions within tasks that require tracking previously-occurring token sequences \citep{wanginterpretability, merullo2024talking, zhangsame}.  

\subsection{Motivation for Current Work}\label{sec:motivation}
However, not all next-token predictions can be made (successfully) by simply tracking and copying previously occurring token motifs. 
One longstanding challenge in natural language processing (NLP) is lexical ambiguity, where a single word points to multiple related (polysemous) or unrelated (homonymous) meanings \citep{navigli2009word, haber2024polysemy}. Next-token completions predicated on ambiguous words require iteratively teasing apart the cued and uncued meanings that are entangled in the ambiguous static token embedding \citep{grindrod2024transformers}. 

Lexical ambiguity is pervasive, with some estimates placing the rate of English-language polysemous words at approximately $80\%$ \citep{rodd2004modelling}. With the majority of words in the English language proving polysemous, and a smaller (but nontrivial) fraction considered homonymous \citep{dautriche2015weaving}, researchers have examined the extent to which Transformer-based LMs \textit{can} tease apart entangled meanings in static embeddings \citep{haber2020word,garcia2021exploring,trott2021raw,gari2021let,haber2021patterns,riviere2025evaluating}. \textbf{To date, however, little work has leveraged ambiguity with the goal of characterizing specialization in attentional patterns during contextualization---or investigated the developmental processes underlying disambiguation performance in the final model}. Understanding how attention heads differentiate meanings across contexts provides a  window into the mechanisms that support contextualization and semantic composition in Transformer LMs.

Investigating the mechanisms underlying model behavior has generally taken one of two prominent forms, involving either the modification of activation patterns and recording the resulting effects on LM logit outputs \citep{wanginterpretability,conmy2023towards}; or the modification of learned static model parameters \citep{nelson2021mathematical,olsson2022context,merullo2024talking,chang2025bigram}. The chief virtue of the latter is the ability to directly edit the information the LM has encoded about the statistics of a linguistic corpus. Of note, the query and key matrices of an attention head dictate which tokens its attention will be allocated to in the preceding context. Specialization in these matrices should heavily contribute to successful contextualization, and indeed, targeted ablations of induction heads does demonstrably impair in-context learning in a range of LMs of differing sizes and architectures \citep{olsson2022context}. 

\section{Phase 1: Identification}\label{sec:phase1}

In Phase 1, we first aimed to characterize the \textit{developmental trajectory} of disambiguation performance over the course of pretraining in the Pythia suite of models (from $14M$ to $12B$). This included questions about the overall shape of this trajectory and the timing of specific ``milestones'' (i.e., \textit{when} marked changes occurred). 

Second, with a focus on $14M$ and $410M$ specifically, we attempted to identify specific attention heads that directed attention from the target ambiguous word (e.g., ``lamb'') to the disambiguating cue (e.g., ``marinated'')---and further, whether the developmental trajectories of these heads overlapped with changes in disambiguation performance. These \textit{candidate disambiguation heads} could then be further stress-tested (Section \ref{sec:phase2}) and ablated (Section \ref{sec:phase3}) to assess both their generalizability and functionality.

\subsection{Dataset}

We used the RAW-C dataset \citep{trott2021raw} as a behavioral probe. RAW-C contains relatedness judgments of ambiguous English words across 672  minimal sentence pairs. Each word appears in four sentences, with two sentences per sense, resulting in six sentence pairs per word (corresponding to four unique Different-Sense pairs, and two unique Same-Sense pairs). Each sentence pair is associated with a mean relatedness judgment, ranging from $1$ (totally unrelated) to $5$ (same sense).

The full dataset contains both ambiguous nouns and ambiguous verbs: nouns are disambiguated solely by a prenominal modifier (e.g., ``marinated lamb''), while verbs are disambiguated solely by a post-verbal clause (e.g., ``broke the promise''). Because Pythia models are auto-regressive and only attend to previous contexts, we used only the noun stimuli (comprising 504 sentence pairs total). Of these, 318 of the ambiguous word senses were classified as polysemous; 186 were classified as homonymous (see \citet{trott2021raw}). 

The sentence \textit{pairs} were used to characterize changes in disambiguation performance, while \textit{individual sentences} were used as inputs for the attention head analysis.

\subsection{Language Models}

To characterize overall disambiguation performance at select checkpoints, we selected the Pythia suite of language models (LMs), a series of English LMs ranging in size from $14M$ to $12B$ parameters, all trained on the same data \citep{biderman2023pythia, van2025polypythias}. All models were run on $20$ checkpoints\footnote{Complete list of checkpoints: [0, 1, 2, 4, 8, 16, 32, 64, 128, 256, 512, 1000, 2000, 5000, 10000, 25000, 50000, 75000, 100000, 143000].}, and $14M$ specifically was run on all available 154 checkpoints. 

We selected LMs $14M$ and $410M$ for analysis of attention head behavior. Pythia-$14M$ was the smallest model and thus a suitable ``model organism'' for developing an experimental protocol. Pythia-$410M$, although far from the biggest model, performed nearly as well as $12B$, rendering it both \textit{performant} and \textit{tractable} as a subject of analysis. For $14M$, we assessed all 154 checkpoints; for $410M$, we assessed $20$ checkpoints. We also assessed the attention head behavior of all models at their final step.

In general, the Pythia suite was well-suited for our research questions in several ways. First, a number of pre-training checkpoints are made publicly available, facilitating analysis of the developmental trajectory of our target behavior. Second, in addition to a ``main'' model release, each model (at each checkpoint) was trained on nine training runs, each initialized with a different random seed. This allows us to evaluate the \textit{robustness} of our empirical results, while controlling for model architecture and training specifications. The primary analyses in this manuscript focus on the ``main'' release of each model. Analyses of random seeds are discussed in Appendix Section \ref{sec:random_seeds}. All models were accessed through the HuggingFace \textit{transformers} library \cite{wolf-etal-2020-transformers}. Models were run either on a Mac laptop (M2, 2022) or on an NVIDIA DGX-H200.

\subsection{Evaluating Disambiguation Performance}

\begin{figure*}[t]
\begin{center}
\includegraphics[width=1\linewidth]{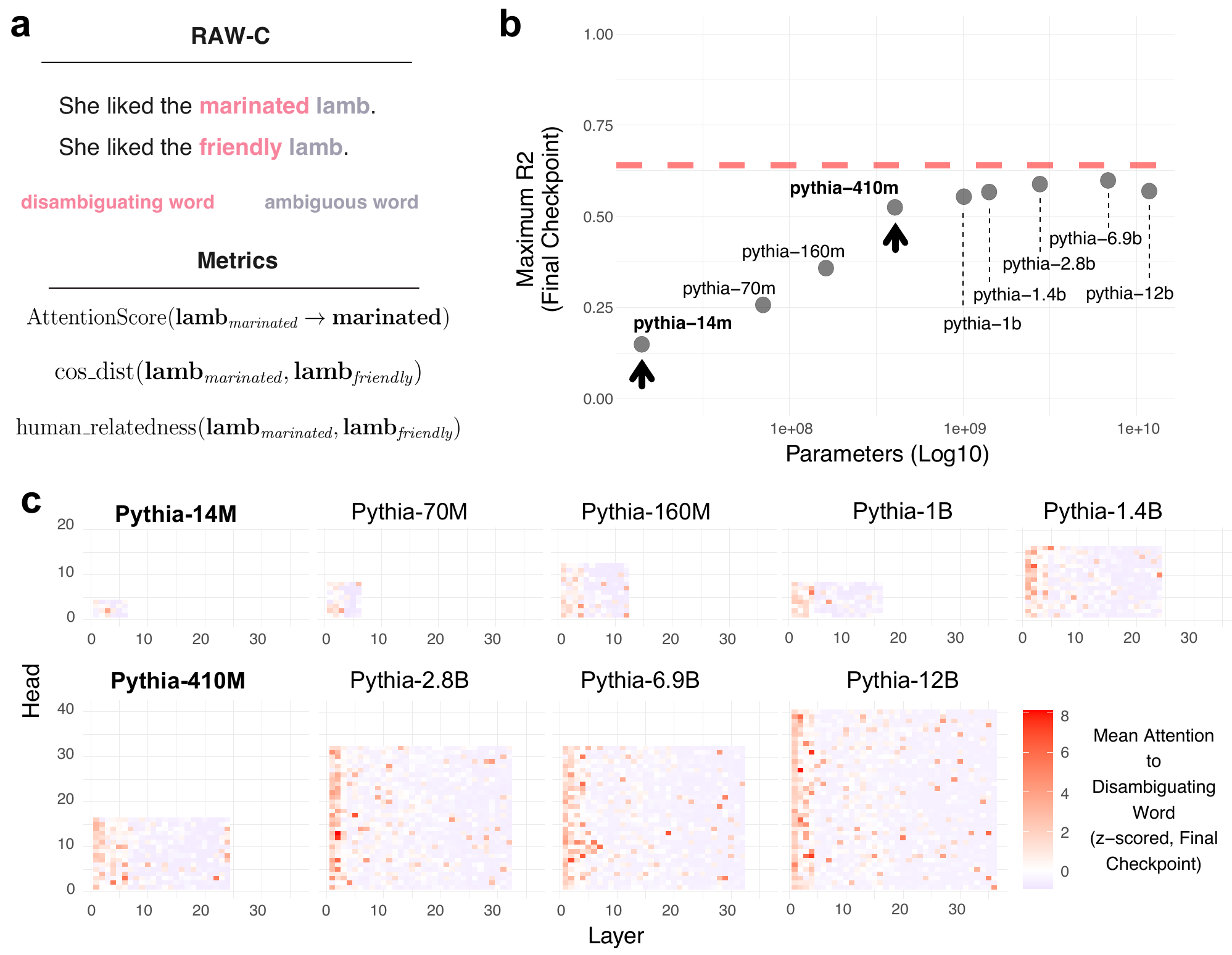} 
\end{center}
\caption{\textbf{Disambiguation performance at final checkpoint for Pythia language models (LMs).} \textbf{(a)} Sample RAW-C sentences, evoking different senses for the target ambiguous word (lamb) with single differing disambiguating word (marinated, friendly). We obtain: the AttentionScore from the ambiguous word to
disambiguating word, per sentence; the cosine distance between contextualized representations for the target ambiguous word across sentences in a pair; and the publicly available human relatedness judgments between the target ambiguous
word across sentences in a pair.  \textbf{(b)} Max $R^2$ obtained from the final checkpoint of nine Pythia LMs, by number of parameters. Arrows mark models of interest, -$14M$ and -$410M$. Horizontal dashed line represents mean human interannotator agreement. \textbf{(c)} Each subpanel shows the head index by layer index for a given LM; warmer colors indicate higher z-scored mean attention scores to disambiguating words, for the final checkpoint of each LM.} \label{fig:fig1-establish-task-metrics}
\end{figure*}

Following past work \citep{nair2020contextualized, trott2021raw, riviere2025evaluating, bojanowski2017enriching, schlechtweg2020semeval}, we calculated the \textit{cosine distance} between the contextualized embeddings of the target word across each sentence pair (e.g., ``She liked the marinated \textit{lamb}'' vs. ``She liked the friendly \textit{lamb}'', \textbf{Figure \ref{fig:fig1-establish-task-metrics}a}). In cases where the target word was tokenized into multiple tokens, we computed the average embedding of those tokens. We repeated this procedure for each layer of each model, such that a given sentence pair was associated with $L$ distance measures for a given model (where $L$ is the number of layers in the model). Then, for each layer $\ell$ of each model, we regressed:

\begin{equation}
    Relatedness \sim Distance_{\ell}
\end{equation}

The resulting $R^2$ measure reflects the proportion of variance in human relatedness judgments explained by the distribution of cosine distances obtained from layer $\ell$. For the primary \textit{developmental analyses} below, we selected the $R^2$ at each checkpoint from the layer that performed best at the \textit{final step}. For example, if Layer $3$ in $14M$ achieved the best $R^2$ at the final step, we analyzed the trajectory of performance at Layer $3$ specifically. Note that we obtained qualitatively similar results using the best $R^2$ from each checkpoint, independent of layer. For instance, when considering all $154$ checkpoints for which $14M$ was assessed, the correlation in $R^2$ trajectories across these measures was $r = 0.96$; moreover, when considering the subset of $20$ checkpoints for which we assessed all models in the Pythia suite, the correlation between these measures ranged from $r = 0.98$ (for $14M$) to $r = 1$ (for 410M, 1B, 1.4B, 2.8B, and 6.9B). 

\subsection{Identifying Candidate Attention Heads}

To isolate candidate attention heads contributing to disambiguation performance, we calculated a \textit{disambiguation score} for each head at each model checkpoint. Specifically, each sentence was tokenized and presented to a given model in isolation. Then, for each head, we obtained the attention score from the target token (e.g., ``lamb'') to the disambiguating cue (e.g., ``marinated''). If either the target token or the disambiguating cue consisted of multiple tokens, we averaged across scores for those tokens. We then performed a linear regression analysis to identify which attention heads showed changes in attention over pretraining that best predicted changes in disambiguation performance.

\subsection{Results}\label{sec:phase1_results}

\begin{figure*}[t]
\begin{center}
\includegraphics[width=0.85\linewidth]{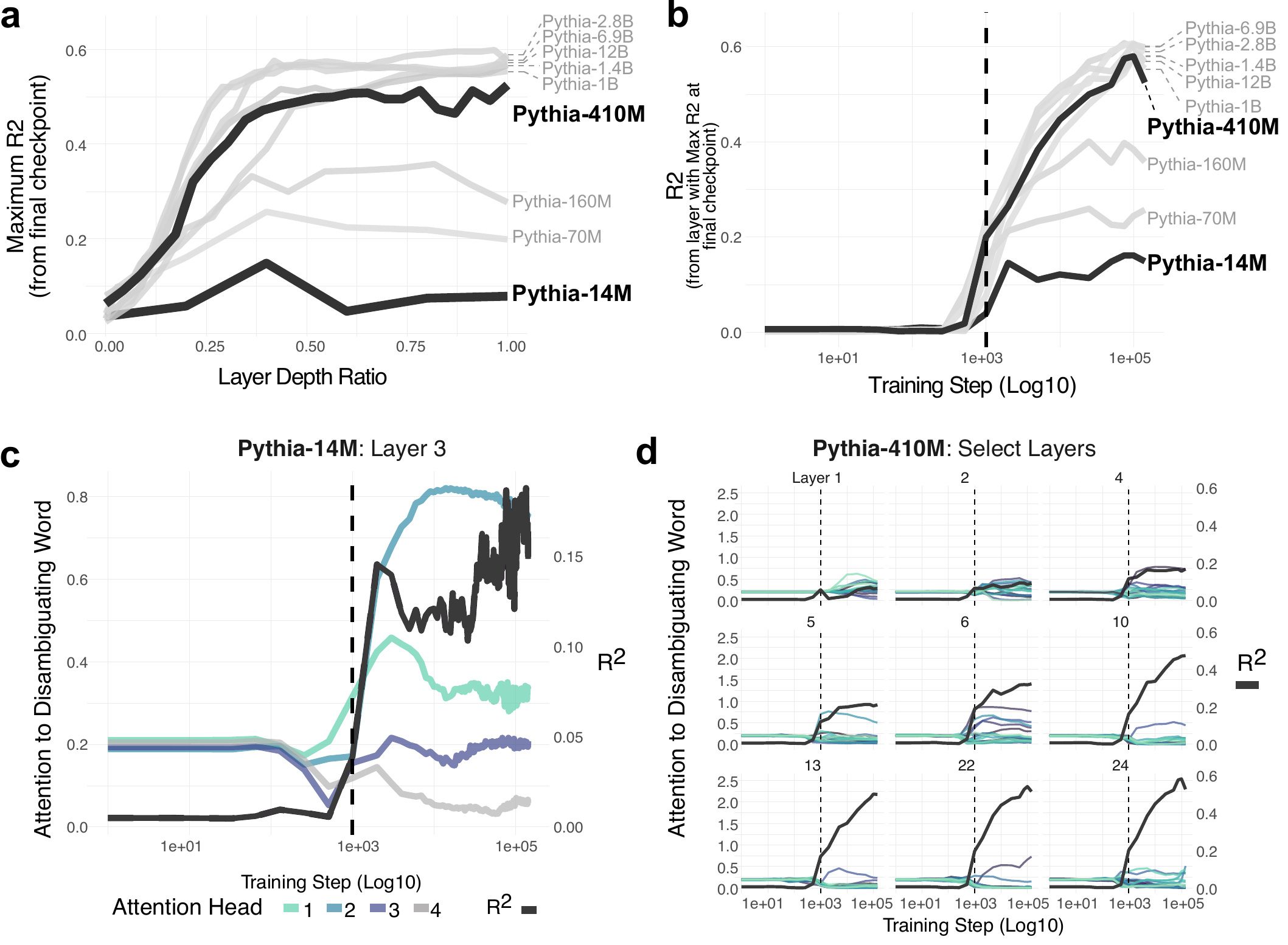} 
\end{center}
\caption{\textbf{Identifying candidate attention heads.} \textbf{(a)} Maximum $R^2$ from the final step, by layer depth (current layer$/$max number of layers), for nine Pythia LMs. \textbf{(b)} ``Developmental'' view of $R^2$, obtained from each training step for nine Pythias. Depicted $R^2$s are from the layer with the max $R^2$ at the final checkpoint (e.g. for $14M$, $R^2$ is from Layer 3; for $410M$, $R^2$ is from Layer 24). \textbf{(c)} ``Developmental'' view of attention to disambiguating word, for all head indices in $14M$'s Layer 3. Superimposed is the $R^2$ from Layer 3. Attention scores for Heads $(3,1)$ and $(3,2)$ covary with disambiguation performance. \textbf{(d)} Same as in \textbf{c}, but for select layers in $410M$. Layers were selected if they contained at least one head whose attention scores rose during training. Superimposed $R^2$ curves were drawn from each Layer depicted. Vertical dashed line in sub-panels \textbf{b-d} mark training step $1000$, corresponding to $2.1M$ tokens seen cumulatively over training.} \label{fig:fig2-identify-candidates}
\end{figure*}

\subsubsection{Disambiguation Performance}

We first quantified the disambiguation performance of each model (as measured by $R^2$) at the \textit{final training step}. As depicted in \textbf{Figure \ref{fig:fig1-establish-task-metrics}b}, larger models performed better than smaller models: $14M$ achieved an $R^2$ of $0.15$, while $6.9B$ (the best-performing model) achieved an $R^2$ of $0.598$. Moreover, there was a significant relationship between model size (Log Number of Parameters) and disambiguation performance ($R^2$) [$\beta = 0.16, SE = 0.02, p < 0.001$]: each order of magnitude increase in model size was associated with a $0.16$ improvement in task performance. Interestingly, however, there appeared to be diminishing returns to scale on this task: $410M$ achieved an $R^2$ of $0.524$, which was $92\%$ of the performance of $12B$ (the largest model tested)---despite having approximately $3\%$ as many parameters. 
No model obtained human-level performance ($0.64$), though $6.9B$ came the closest. Notably, the maximum $R2$ was not always derived from the final layer's representations, as in the case of Pythia-$14M$, whose maximum $R2$ came from Layer 3 (\textbf{Figure\ref{fig:fig2-identify-candidates}a}).

We then examined the developmental trajectory of disambiguation performance. We found clear evidence of discontinuities in $R^2$ throughout pretraining, with marked shifts in performance at step $1000$ and step $2000$---corresponding to $2.1$B and $4.2$B tokens seen, respectively (\textbf{Figure\ref{fig:fig2-identify-candidates}b}). Focusing specifically on $14M$ (for which all 154 checkpoints were assessed), the model had achieved close to its final-step performance ($0.15$) by step 2000 ($4.2$B tokens). This trajectory is consistent with past work documenting sudden ``phase shifts'' in model performance \citep{hu2023latent, chensudden}; moreover, it is striking that close-to-maximal performance was achieved quite early in pre-training---approximately $1.4\%$ of the way through (\textbf{Figure\ref{fig:fig2-identify-candidates}b}).\footnote{Interestingly, we observed a subsequent ``dip'' in performance using Layer $3$ representations between steps $5000$ and $50000$. Using the $R^2$ from the \textit{best-performing layer} at each step instead resulted in a somewhat smoother (though qualitatively similar) trend.} A linear regression predicting $R^2$ from the Log Training Step found a significant relationship between the two variables $[\beta = 0.04, SE = 0.001, p < .001]$, with Log Training Step explaining about $85\%$ of the variance over time in $R^2$.

In contrast to $14M$, larger models appeared to continue improving well past these early steps---perhaps suggesting that larger model capacity was better able to reap the benefits of the training data \citep{DBLP:journals/corr/abs-2001-08361, 10.5555/3600270.3602446}.

\subsubsection{Identifying Candidate Attention Heads}

We then identified candidate attention heads contributing to these changes in performance. Heads displayed considerable variance in the degree to which they attended to the disambiguating cue at the final step (\textbf{Figure \ref{fig:fig1-establish-task-metrics}c}).

In Pythia-$14M$, the strongest final-step attention to the disambiguating word came from Head $2$ in Layer $3$ (hereafter Head $(3, 2)$); this head's average attention score ($0.76$) was approximately $3.7$ standard deviations over the mean attention to the disambiguating word in Pythia-$14M$ heads. Further, as seen in \textbf{Figure \ref{fig:fig2-identify-candidates}c}, two heads in layer $3$---$(3, 1)$ and $(3, 2)$---exhibited patterns of attention that were (relatively) aligned temporally with changes in disambiguation performance (see Appendix \textbf{Figure \ref{fig:appendix-fig2ccompanion}} for a view of all layers). This was confirmed quantitatively by constructing a series of linear models regressing $R^2$ over training against the attention score $\alpha$ from a given head $h$ over training ($p$-values were corrected for multiple comparisons using a false-discovery rate correction procedure). Strong positive relationships were obtained for Head $(3, 2)$ and Head $(3, 1)$.

Unsurprisingly, $410M$ had a larger number of \textit{candidate heads}, i.e., those with an unusually high mean attention to the disambiguating word. As evident in \textbf{Figure \ref{fig:fig1-establish-task-metrics}c} and \textbf{Figure \ref{fig:fig2-identify-candidates}d}, these heads appeared in a variety of layers. A majority appeared in relatively \textit{early} layers (e.g., layers $1-4$), with some appearing in the final layers (i.e., layers $22-24$; see Appendix \textbf{Figure \ref{fig:appendix-fig2dcompanion}} for a view of all layers). As with $14M$, we constructed a series of linear models regressing $R^2$ over training  against the attention score $\alpha$ from a given head $h$ over training (again correcting for multiple comparisons). The strongest positive coefficients were generally obtained in Layer $1$, though we also observed robust relationships in later layers.

\section{Phase 2: Stress Testing to Define Attentional Scope}\label{sec:phase2}

In Phase 1, we found marked \textit{phase shifts} in disambiguation performance that coincided with changes in the behavior of select attention heads, which systematically directed attention from the target word (e.g., ``lamb'') to the disambiguating cue (e.g., ``marinated''). After \textit{identifying} these candidate attention heads, we sought to assess the robustness and selectivity of their attention patterns for disambiguating words. RAW-C sentence stimuli always contain the key disambiguating modifier immediately preceding the target ambiguous word. Our goal was to evaluate the extent to which this behavior was robust to a range of stimulus perturbations (see Section \ref{subsec:stimuli_manipulations}), as well to compare the behavior to simpler, ``lower-level'' functions such as ``1-back attention''.

\subsection{Stimuli Manipulations}\label{subsec:stimuli_manipulations}

We devised three ``control'' tasks to \textit{stress-test} \citep{shapira-etal-2024-clever,naik-etal-2018-stress} the putative disambiguation circuits. 

In the \textbf{1-back analysis}, we measured the average attention directed by each head from each token in a sentence to the immediately preceding token \citep{clark2019does}. We then performed a \textit{subtraction analysis}, which allowed us to determine whether attention to the immediately preceding disambiguating cue was particularly strong, i.e., over and above 1-back tokens in general.

With \textbf{Positional modification}, we modified the \textit{position} of the disambiguating cue relative to the target noun by adding a semantically bleached phrase between the disambiguating cue and the target noun (e.g., ``{tense/gaseous} \textit{kind of} atmosphere''). Then, as in Section \ref{sec:phase1}), we calculated the attention for each head from the target noun to the disambiguating cue. This allowed us to determine whether the same heads still attended to the nominal modifier even when it was separated from the target noun ($N = 310$ sentences).

Finally, with \textbf{Part-of-speech modification}, we modified the RAW-C sentences \citep{trott2021raw} such that ambiguous nouns were now disambiguated by a \textit{verb} (e.g., ``He \textit{polished/filed} the case''); we also included sentences with ambiguous verbs, which were disambiguated by a noun phrase in the subject position (e.g., ``The \textit{glass/promise} was broken''). Then, we calculated the attention for each head from the target word to the disambiguating cue. This allowed us to determine whether the same attention heads directed attention to disambiguating cues regardless of their part-of-speech ($N = 360$ sentences).

The analyses described above were carried out across all checkpoints of the Pythia-$14M$ model, as well as the $20$ checkpoints of the Pythia-$410M$ model investigated in Section \ref{sec:phase1}.

\subsection{Results}

\subsubsection{1-back attention}\label{sec:phase2-1back}

Multiple heads in both models emerged as candidates for ``1-back heads'', some of which overlapped with the candidate heads identified in Phase 1. For example, in $14M$, Head $(3, 2)$ underwent a developmental trajectory that strikingly resembled the patterns described in Section \ref{sec:phase1}, i.e., changes in attention emerged at roughly $2000$ steps. We then asked whether the heads identified in Section \ref{sec:phase1} showed particularly strong attention to the disambiguating word, \textit{above and beyond} their 1-back attention more generally. For each sentence and each head at each checkpoint, we subtracted the 1-back attention from that head's attention to the disambiguating word. Finally, we performed a one-tailed paired $t$-test at each checkpoint for each head (correcting for multiple comparisons using a false-discovery rate (FDR) procedure across all checkpoints and heads). This procedure revealed several heads with significantly higher attention to disambiguating words. 

In $14M$, these heads included $(3, 1)$ and Head $(3, 2)$ (see \textbf{Figure \ref{ref:fig3-phase2-stress}a}). In $410M$, $11$ heads survived the subtraction analysis, including $(1, 11), (1, 14), (6, 5),$ and $(23, 8)$. These heads may be specialized for attention to prenominal modifiers above and beyond 1-back attention. 

\subsubsection{Positional modification}

We then asked whether attentional patterns were robust to the relative position of the prenominal modifier (e.g., ``friendly sort of lamb''). 

\begin{figure*}

\includegraphics[width=1\linewidth]{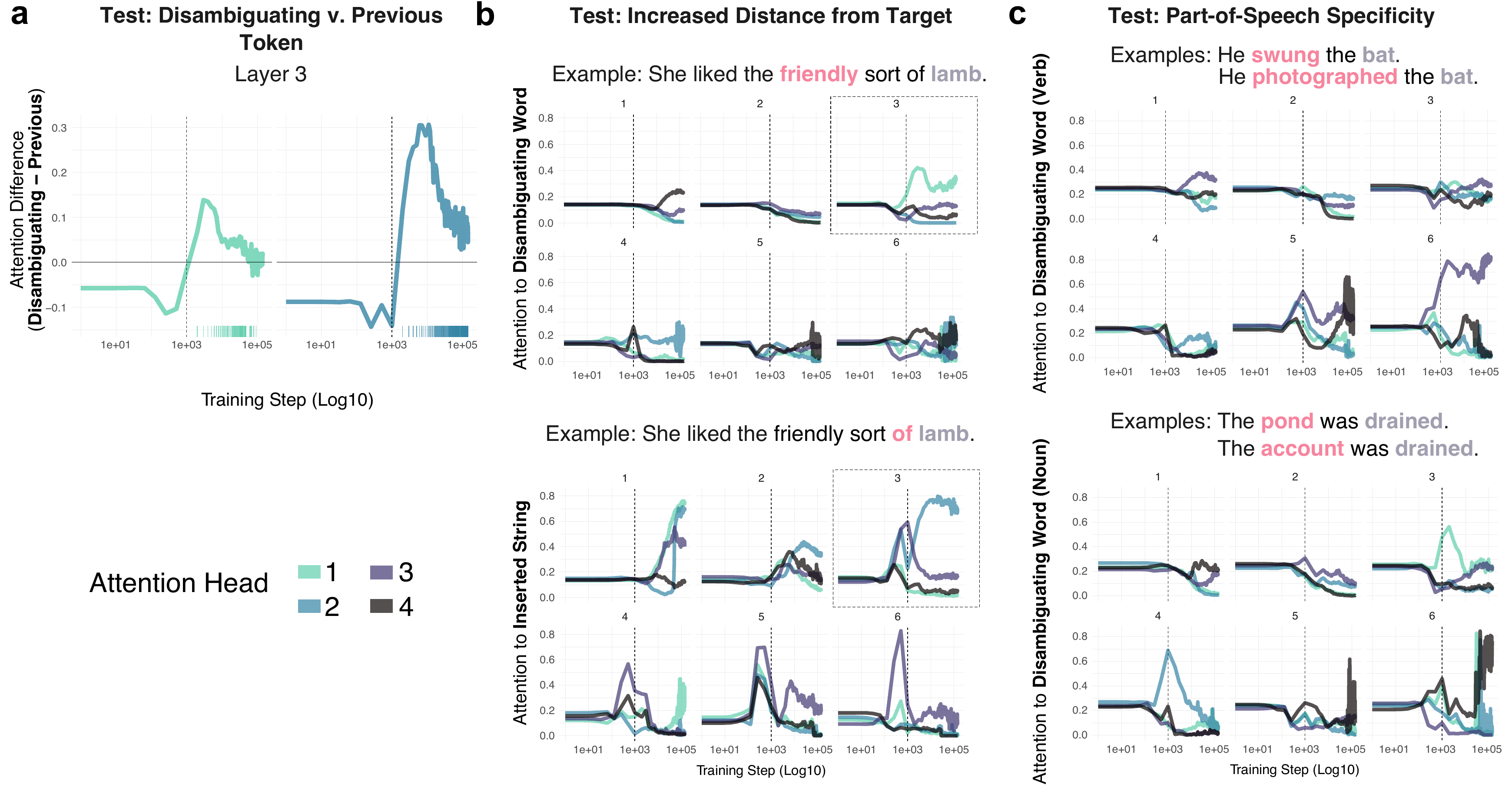}
\caption{\textbf{Stress testing Pythia-14M candidate heads' attention to disambiguating word.} Attention Head color coding scheme applies to all panels. \textbf{(a)} Layer 3 difference in average attention scores for disambiguating word against that of all 1-back tokens, over pre-training. Ticks indicate training steps with significant difference in attention scores (p < 0.05), adjusted for multiple comparisons. Only Head (3,2) remains significant at the final training checkpoint.  \textbf{(b)} Each sub-panel corresponds to a different layer. (\textit{top}) Attention to disambiguating word when it is separated from the target ambiguous word via inserted string. (\textit{bottom}) Attention to last token of inserted string. The square surrounding Layer 3 highlights the attentional robustness of at least one of the two candidate heads. \textbf{(c)} Attention to disambiguating word when its part-of-speech changes to a verb (\textit{top}) or a noun (\textit{bottom}). }\label{ref:fig3-phase2-stress}
\end{figure*}

In $14M$, only a single attention head---Head $(3, 1)$---appeared to be robust to this modification, showing a very similar developmental trajectory as it did for the original stimuli (\textbf{Figure \ref{ref:fig3-phase2-stress}b, \textit{top}}). In contrast, Head $(3, 2)$ (previously identified as attending strongly to the prenominal modifier) directed attention primarily to the token immediately preceding the target (e.g., ``friendly sort \textbf{of} lamb''; \textbf{Figure \ref{ref:fig3-phase2-stress}b, \textit{bottom}}).\footnote{Interestingly, this stimulus modification brought to the fore multiple heads in Layers 1, 2, and 4 that attended substantially to the last token of the inserted string, although they were previously inattentive to the disambiguating word when it immediately preceded the target ambiguous word in original stimuli. Given that this token was typically a high-frequency preposition (e.g., ``of''), it is possible these heads are sensitive to lexical properties, e.g., unigram frequency.}

In $410M$, a number of heads consistently directed attention to the disambiguating cue at the final step---and also showed strikingly similar developmental trajectories as they did for the original stimuli. This included a number of previously identified from Layer $1$, such as $(1, 7)$ and $(1, 14)$, but also included heads from later layers, such as $(6, 4)$ and $(24, 8)$.

\subsubsection{Part-of-speech modification}

Finally, we asked which attention heads, if any, attended to disambiguating words regardless of their part-of-speech. In $14M$, we found virtually orthogonal sets of attention heads (mostly in layer $6$) that directed attention to disambiguating \textit{verbs} (e.g., ``He \textbf{swung} the \textit{bat}'') vs. disambiguating \textit{nouns} (e.g., ``The \textbf{pond} was \textit{drained}'') (\textbf{Figure \ref{ref:fig3-phase2-stress}c}). Notably, neither set of heads overlapped substantively with the original heads identified in Section \ref{sec:phase1}, with the possible exception of Head $(3, 1)$ attending to disambiguating nouns at approximately step $1000$. 

In $410M$, on the other hand, there were select heads that consistently directed attention to the disambiguating cue, regardless of its part-of-speech. Although these heads appeared to show some preference for certain parts-of-speech (e.g., nouns over verbs), the \textit{minimum} attention directed to disambiguating cues was systematically higher than other heads independent of the cue's part-of-speech. Candidate heads identified in this process included $(1, 7), (1, 14), (4, 7)$, and $(6, 10)$.\footnote{There were also select heads with a strong \textit{part-of-speech bias}, as measured by the difference in average attention directed towards disambiguating nouns vs. verbs. Noun-preferring heads included $(6, 5)$ and $(23, 11)$, and verb-preferring heads included $(4, 4)$ and $(11, 2)$.}

\begin{figure}[t]
 \includegraphics[width=1\linewidth]{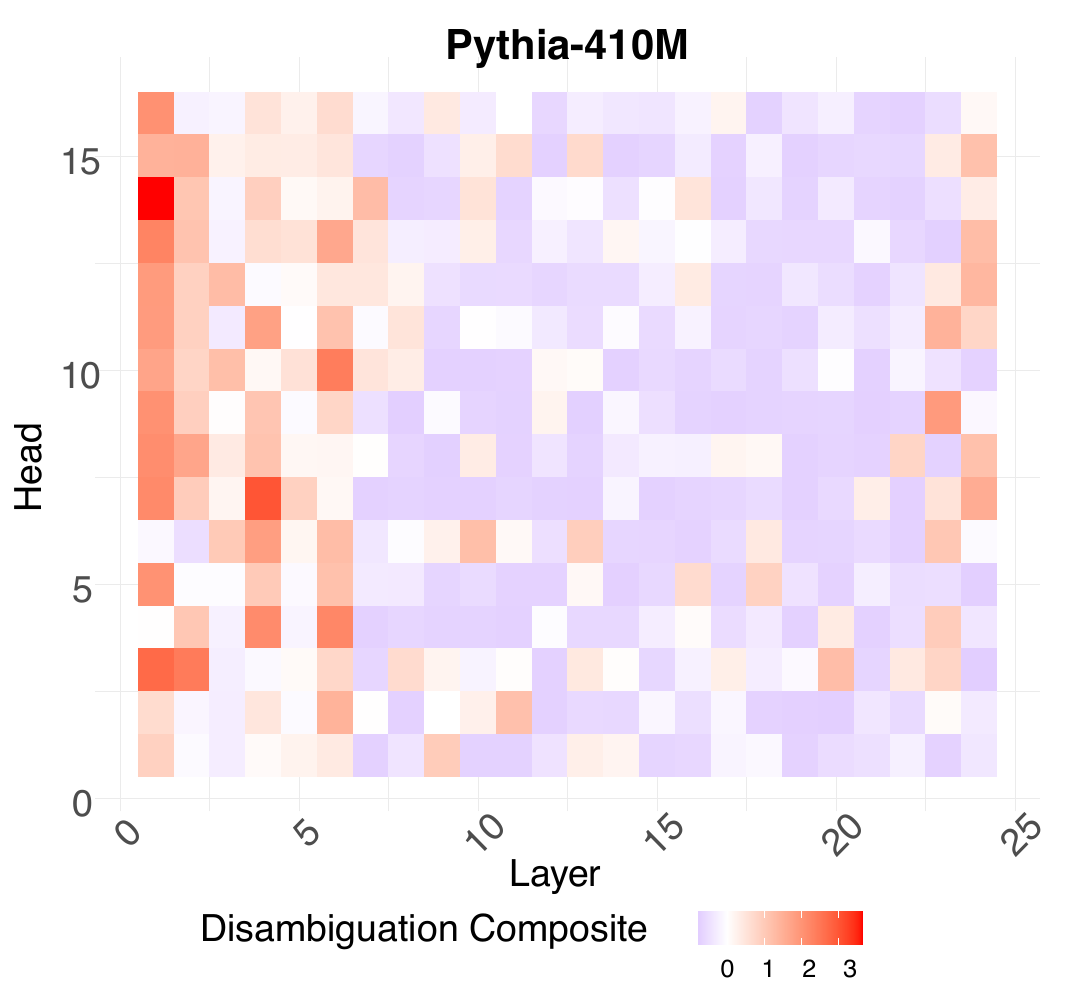} 
 \caption{\textbf{Stress testing Pythia-$410M$.} Warmer colors indicate larger Disambiguation Composite scores for Pythia-410M heads and layers. Larger scores reflect greater attentional robustness to the disambiguating word despite stimulus perturbations, and greater degree of attention covariance with disambiguation performance throughout pre-training.}\label{fig:pythia-410m-disamb-composite}
\end{figure}

\subsubsection{Constructing a ``Disambiguation Index'' for $410M$ Heads}\label{subsec:index}

To facilitate interpretation for stress-test results in a model containing a much larger number of attention heads, we constructed a composite score for each head in $410M$, reflecting a given head's robustness to stimulus perturbations and its attentional covariance with performance over pre-training. Specifically, this score is constructed by taking the average of five $z$-scored variables: the coefficient relating changes in attention to changes in $R^2$; the average final-step attention to disambiguating nouns; the average final-step attention to disambiguating verbs; the $t$-statistic resulting from the $1$-back subtraction analysis; and the average final-step attention to disambiguating modifiers in the ``sort of'' analysis. A higher score on this index (e.g., $>2$) indicated the extent to which a given head consistently (i.e., \textit{robustly}) attended to disambiguating cues across various stimulus perturbations and the extent to which this covaried with changes in disambiguation performance. This composite score is depicted for each head in \textbf{Figure \ref{fig:pythia-410m-disamb-composite}}. Most heads received low scores, while only a handful received very high scores. The six heads with the largest score on this index included $(1, 14), (4, 7), (1, 3), (2, 3), (6, 10)$, and $(1, 13)$.

\subsubsection{Summary of Stress-Testing Results}

Our primary goal in Phase 2 was \textit{stress-testing} the heads identified in Phase 1: namely, how \textit{specialized} are their putative functions, and how \textit{robust} is their behavior to stimulus perturbations? 

The results for $14M$ were mixed: while select candidate heads may be specialized for more than relatively simple mechanisms like 1-back attention, their behavior did not generalize across sentence frames or parts of speech, suggesting that they are not ``generalized disambiguation heads'' per se---rather, they may \textit{participate} in disambiguation for nouns specifically. In contrast, $410M$ contained select heads whose attentional patterns were surprisingly robust to stress-testing, as revealed by particularly high scores on the \textit{disambiguation index} (Section \ref{subsec:index}; \textbf{Figure \ref{fig:pythia-410m-disamb-composite}}), suggesting their \textit{functional scope} was considerably broader---or at least more abstract---than the heads identified in $14M$. 

Crucially, however, Phase 2 investigated only the \textit{attentional patterns} of these heads; in Phase 3, we examine their putative \textit{causal contributions} to disambiguation.

\begin{figure*}[t]
\begin{center}
 \includegraphics[width=0.80\linewidth]{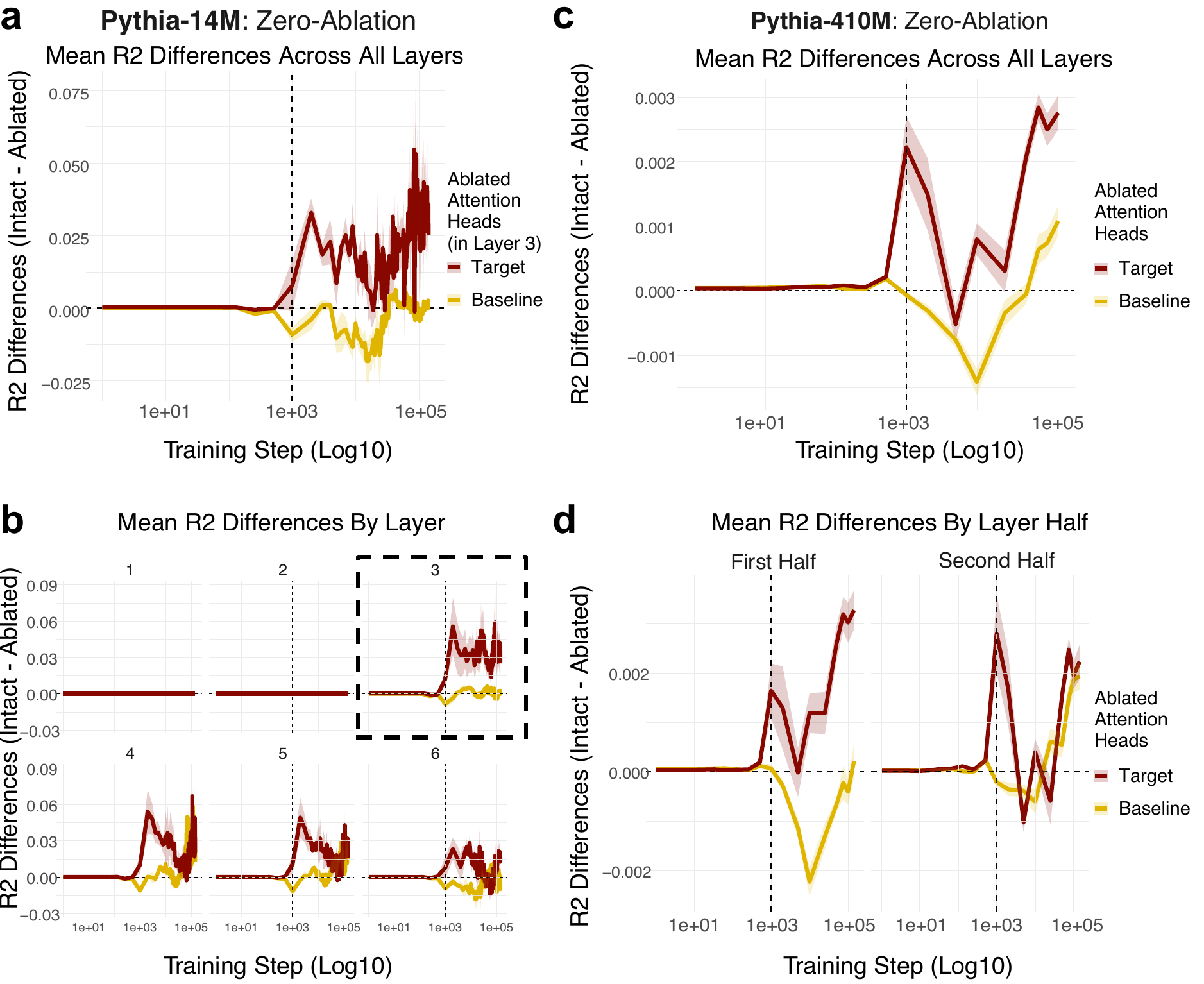} 
\end{center}
\caption{\textbf{Target head ablations decrease disambiguation performance relative to intact models.} \textbf{(a)} Mean difference in $R^2$ across all layers and all combinations of Pythia-$14M$'s target head ablations, for all training steps. Values $>0$ indicate that the intact model's $R^2$ exceeded that of the ablated model's $R^2$, reflecting causal effect of ablation. Target manipulations refer to zero-ablations of previously-identified Layer 3 heads. Baseline manipulations refer to zero-ablations of Layer 3 heads whose attention to disambiguating words fail to increase with disambiguation performance. \textbf{(b)} Same as in \textbf{a}, but parcelled out by layer, to illustrate localization of ablation effects, which remain robust throughout training in Layer 3. Dashed square marks the only layer (Layer 3) to suffer head ablations. \textbf{(c)} Same as in \textbf{a}, but for Pythia-$410M$. Target head ablations causally decrease model performance. \textbf{(d)} Same as in \textbf{c}, but parcelled out by early versus late layers, to illustrate the selectivity of target-head ablation to earlier layer representations. By the end of training, the effects of target-head ablations are indistinguishable from those of baseline-head ablations.}\label{fig:fig5_zero-ablation}
\end{figure*}

\section{Phase 3: Causal Analysis}\label{sec:phase3}

To determine the extent to which candidate heads' attention patterns are \textit{necessary} for Pythia-$14M$'s and Pythia-$410m$'s disambiguation performance at different pre-training stages, we carried out a series of targeted ablations at each model checkpoint. We then asked how ablating the target heads identified in Phase 1 and Phase 2 affected the model's performance on the disambiguation task (relative to ablating a set of ``control'' heads). 

\subsection{QK Matrix Manipulations}

Attention heads in the Transformer architecture consist of matrices of learnable parameters. A given head's Query ($\textbf{W$_Q$}$) and Key ($\textbf{W$_K$}$) matrices specifically direct attention to select tokens in the input sequence. To directly intervene on our candidate head's attention patterns, we manipulated the values of these two matrices.

In $14M$, we selected Head $(3,1)$, Head $(3,2)$, and the combination of both heads. To intervene specifically on our candidate attention heads, Heads $(3,1)$ and $(3,2)$, we directly modified the weights in their query-key (QK) matrices. In the \textbf{Zero-Ablation} condition, we set the target head's $\textbf{W$_Q$}$ and $\textbf{W$_K$}$ to equally-sized matrices of zeros. In the \textbf{Step1-Copy-Ablation} condition, we set the target head's $\textbf{W$_Q$}$ and $\textbf{W$_K$}$ matrices to the parameter values they held at the first pre-training checkpoint (i.e., step 1); this latter ablation ensured that the target head(s) is(are) still participating in transforming embeddings, but with suboptimal parameters. We implemented both ablation types for Heads $(3,1)$ and $(3,2)$ on their own as well as concurrently, at each training step beginning with step $1$. To control for the possibility that \textit{any} modifications at this layer would result in a reduction in performance, we implemented a set of \textit{baseline} conditions, in which we ablated two other heads from the same layer: $(3, 3)$ and $(3, 4)$---again, either on their own or together.

We followed the same set of procedures for select heads in $410M$. In this case, we selected the six heads with the top \textit{disambiguation index}, as identified in Section \ref{subsec:index}; heads were shown to be relatively robust to stimulus perturbations.\footnote{Note that in theory, the heads with the most robust behavior need not be the heads most involved in a \textit{particular} stimulus configuration, i.e., the one assessed in the original RAW-C stimuli; nevertheless, we selected these heads because they were the most likely candidates for serving the purpose of ``disambiguation'' across a number of contexts.} These heads included: $(1, 14), (4, 7), (1, 3), (2, 3), (6, 10), (1, 13)$. As baseline heads, we selected a \textit{matched control} for each target head from the same layer but with a lower disambiguation index. This ensured that the controls were matched for layer (as ablating heads in distinct layers could affect measured $R^2$ using representations from, say, the final layer).

\subsection{Procedure}

We followed the same procedure described in Phase 1 (Section \ref{sec:phase1}) for measuring the disambiguation performance (i.e., $R^2$) of each ablated model at each checkpoint. We then calculated the difference between each ablated model's $R^2$ and the intact model's $R^2$ at an equivalent checkpoint ($\Delta R^2$), as well as the ratio between the ablated and intact models' $R^2$ values (``fraction of $R^2_{Intact}$''). 

\subsection{Results}

As depicted in \textbf{Figure \ref{fig:fig5_zero-ablation}a\&c}, ablating the target heads resulted in impaired performance in both $14M$ and $410M$ on average across layers. This effect was systematically larger when ablating target heads relative to ablating the baseline heads, and the \textit{developmental trajectory} of performance reductions was reflective of the trajectories reported in Phase 1 and Phase 2: i.e., clear divergences first emerge at approximately $1000$ steps. 

Notably, the effect of ablating target heads depended to some extent on which layers were used to extract contextualized representations. In $14M$, the strongest effects were observed in Layer $3$ (\textbf{Figure} \textbf{\ref{fig:fig5_zero-ablation}b}), which was also the maximally-performing layer (as identified in \textbf{Figure \ref{fig:fig2-identify-candidates}a}); in $410M$, the impact of target-head ablation was most pronounced in the first twelve layers, and was not distinguishable from baseline-head ablations in the final layers (\textbf{Figure \ref{fig:fig5_zero-ablation}d}) other than training step $1000$. Similarly, ablating the different target heads in $14M$ produced different effects at this model's different layers and checkpoints: ablating Head $(3,1)$ impaired performance earlier in pretraining, while ablating Head $(3,2)$ hurt performance in later stages. (Ablating \textit{both} resulted in more consistent reductions in performance throughout pre-training; see Appendix \textbf{Figure \ref{fig:appendix-fig5companion-14M-individual-target-heads}}.) These differences mirror the different developmental trajectories of the two heads identified in Section \ref{sec:phase1_results} (see also \textbf{Figure \ref{fig:fig2-identify-candidates}c} and Appendix \textbf{Figure \ref{fig:appendix-fig2ccompanion}}). This question of whether different heads matter to different degrees at different timepoints is explored in more detail in the General Discussion (Section \ref{subsec:baton}).

To quantify the impact of each ablation type, we built a series of linear regression models for both $14M$ and $410M$ predicting $\Delta R^2$ or fraction of $R^2_{Intact}$ as a function of \textsc{condition} (Baseline vs. Target) and \textsc{Log Training Step}. We conducted this analysis both for representations from the optimal layer (Layer 3 in $14M$, Layer 24 in $410M$) and also averaging performance reductions across layers (as in \textbf{Figure \ref{fig:fig5_zero-ablation}a, b}).\footnote{Qualitatively similar results were obtained using the raw training step number. Further, allowing an interaction between \textsc{condition} and \textsc{log training step} suggested that the effect of the Target ablations increased for later steps.} 

In $14M$, a significant coefficient was found for \textsc{condition} for each ablation type and each dependent variable, suggesting that variability in both indices was attributable to ablating the target heads. Specifically, zero-ablating the Target heads in $14M$ was associated with a larger $\Delta R^2$ $[\beta = 0.03, SE = 0.001, p < .001]$ and smaller fraction of $R^2_{Intact}$ $[\beta = -0.21, SE = 0.007, p < .001]$ (\textbf{Figure \ref{fig:fig5_zero-ablation}a,c}). The step1-copy ablations were also associated with a larger $\Delta R^2$ $[\beta = 0.05, SE = 0.002, p < .001]$ and smaller fraction of $R^2_{Intact}$ $[\beta = -0.33, SE = 0.01, p < .001]$ (Appendix \textbf{Figure \ref{fig:appendix-fig5companion}a,c}). Qualitatively identical results were found when considering average performance across layers. Put another way, ablating the target heads resulted in measurably larger \textit{reductions} in disambiguation performance than did ablating the baseline heads.

In $410M$, we found a systematic (though small) effect of ablating Target heads on \textit{average} performance across layers, though not on the best-performing layer (see also Figure \textbf{\ref{fig:fig5_zero-ablation}d}). When considering average performance across layers, zero ablations were associated with a larger $\Delta R^2$ $[\beta = 0.001, SE = 0.0002, p = 0.001]$; the effect on $R^2_{Intact}$ was only marginally significant $[\beta = -0.007, SE = 0.004, p = 0.05]$. The step1-copy ablations were also associated with a larger $\Delta R^2$ $[\beta = 0.001, SE = 0.0002, p = 0.003]$ and a smaller fraction of $R^2_{Intact}$ $[\beta = -0.004, SE = 0.001, p < .001]$. These effects were both very small in absolute terms and also absent in the best-performing layer (layer $24$). Together, this suggests that the function of each individual head may overlap, consistent with past work on redundancy in attention heads \citep{michel2019sixteen}; and further, that the impact of ablating target vs. baseline heads in earlier layers was indistinguishable by the final layers.

\section{General Discussion}\label{sec:discussion}

In this work, we report a critical \textit{inflection point} in disambiguation performance of both Pythia-$14M$ and Pythia-$410M$ over the course of pre-training, which coincides with increased attention to the disambiguating word from a subset of attention heads (Section \ref{sec:phase1}); changes in the behavior of certain attention heads account for as much as $77\%$ of the variance in changes in disambiguation performance throughout pre-training. In both models, we observed heads that attended preferentially to the disambiguating cue above and beyond a general preference for 1-back tokens (Section \ref{sec:phase2-1back}). Further, in $410M$, we identified select heads that were also robust to positional and part-of-speech manipulations (Section \ref{subsec:index}), though $14M$ contained no such heads (Section \ref{sec:phase2}). Finally, the ablation analyses point to a clear (and relatively large) \textit{causal role} in disambiguation for the candidate heads in $14M$. The effect of ablating individual heads in $410M$ was weaker (and more pronounced in early layers), though this is less surprising given that $410M$ had 16x as many heads as $14M$ overall (Section \ref{sec:phase3}).

\subsection{On Development and ``Passing the Baton''}\label{subsec:baton}

These results join a growing body of work adopting a ``developmental'' perspective on the mechanisms underlying LM behaviors \citep{chensudden, olsson2022context, van2025polypythias}. As in the study of humans \citep{de2013twelve}, an ontogenetic approach offers unique benefits: in this case, it enables us to identify inflection points in the emergence of capabilities, yielding insights into prerequisite behaviors the LM must achieve to display the capability, and allowing us to link the capability to fine-grained mechanisms.

Here, the results indicate that the representations and mechanisms crucial to contextualizing ambiguous nouns with their modifiers develop relatively early in pre-training. Further, the checkpoints identified align to some degree with previous work on the development of other mechanisms across model scales and random initializations \citep{tigges2024llm, van2025polypythias, olsson2022context, trott2025toward}. Future work could investigate the precise changes in weight matrices that subserve these developments and potentially identify the biases in initial parameterization that lead to different patterns of head ``specialization'' across random seeds (see Appendix \textbf{Figure \ref{fig:generalizability}}). This work could also investigate the role of other model components in disambiguation, such as the value matrices or the residual stream.

We also identified intriguing qualitative changes in attention head behavior (and possible interactions) over the course of pre-training. For instance, some heads played a stronger role earlier in pre-training, then appeared to lose influence as training proceeded. We preliminarily term this phenomenon ``passing the baton'', as other heads (in some cases, heads in the same layer) come to encode the relevant information over the course of pre-training. Similarly, ablations of heads in early layers affected the quality of downstream representations to different degrees at different points in pre-training, potentially reflecting the emergence of alternative routes to transmitting the relevant information. These phenomena are themselves ripe for future investigation, and are only discoverable by adopting a developmental perspective.

\subsection{On Assessing Functional Scope}

Identifying the ``function'' of a circuit (biological or artificial) is notoriously challenging \citep{haklay2025position}. Here, the results of our stress-testing point to qualitative differences in the robustness (and redundancy) of attention head behaviors across $14M$ vs. $410M$.

Specifically, target heads in $410M$ were considerably more \textit{robust} to manipulations of position or part-of-speech than heads in $14M$. One speculative explanation for this difference is that the larger parameter count of $410M$ grants more opportunities for \textit{functional abstraction}. While heads in $14M$ perform operations that ultimately subserve the high-level task we call ``disambiguation'', they may in fact be specialized for lower-level functions---in contrast, heads in $410M$ may be better candidates for generalized disambiguation heads. Notably, this conclusion depends on rigorous stress-testing, i.e., assessing the robustness of a model component's behavior to different perturbations. 

At the same time, ablating heads in $14M$ led to larger reductions in $R^2$ (about 30x) than in $410M$. Again, one explanation for this is the difference in model size: $410M$ has 16x as many heads as $14M$, raising the possibility of some overlap across those heads' functions. From this perspective, the fact that ablating a single head (out of 384) results in any reduction in performance could be seen as surprising. Indeed, there is considerable evidence for \textit{redundancy} in attention head functions in transformer language models \citep{michel2019sixteen, bian-etal-2021-attention, he2024matters, kovaleva-etal-2019-revealing}.

We also note that while we did assess the behavioral robustness of attention heads in both models (see Section \ref{sec:phase2}), we only assessed their functional involvement in disambiguating the original RAW-C stimuli (i.e., the sentence pairs for which we actually had relatedness judgments). Future work could build on the RAW-C dataset and collect human judgments for the modified stimuli as well. Another open question is whether the heads identified in each model are actually \textit{selective} for ambiguous target words in particular, or whether they participate in contextualizing any target word (ambiguous or otherwise). Future research could ask whether the behavior of these heads is robust to the status of the target word, e.g., whether it is ambiguous (``marinated \textit{lamb}'') or unambiguous (``marinated \textit{pork}''). 

\subsection{The Search for Generalization}

The degree to which we are licensed to draw inferences about a wider class of language models from any one pattern of results is an important question to consider. 

Following \citet{trott2025toward} and \citet{fehlauer2025convergence}, the current work does \textit{not} take for granted that the disambiguation performance and/or the corresponding attention specialization patterns obtained in a single model instance (e.g. the default Pythia-$14M$) should necessarily replicate even within other instances of the same model architecture, pretrained on an identical sequence of token batches, for the same number of pretraining steps, but initialized with a different set of randomly selected parameters.

Our work engages with these generalizability considerations directly by (a) replicating experiments with the available Pythia-$14M$'s random seeds (\textbf{Figure \ref{fig:generalizability}; see Appendix Section \ref{sec:random_seeds}}), and (b) replicating the full set of interpretability analyses in a model an order of magnitude larger, Pythia-$410M$, after having assessed disambiguation performance across the entirety of the Pythia suite (from $14M$ to $12B$ parameters). Future work could pursue additional questions of generalizability by carrying out the full set of interpretability analyses on other large LMs in the Pythia suite beyond $14M$ and $410M$, or expand this work to other model families with publicly available checkpoints, e.g., OLMo 2 \citep{olmo20252olmo2furious}. Moreover, questions regarding the generalizability of attentional specialization patterns in the context of \textit{different languages} could leverage the growing body of ambiguity datasets in languages other than English \citep{baldissin-etal-2022-diawug, gari2021let, schlechtweg-etal-2024-durel, abuin2025assessing, riviere2025evaluating}. 

Our attempts to explore generalization across random seeds of Pythia-$14M$ revealed an intriguing pattern of results: while we observed remarkable robustness in terms of \textit{developmental} trajectories of \textit{disambiguation performance} across seeds, there was marked (and systematic) variance in which \textit{layer} hosted the key attention heads (see Appendix \ref{sec:random_seeds}). This suggests that any given capability may be subserved by heterogenous mechanistic solutions---even in the same LM architecture, trained on the same sequence of tokens and following the same training objectives and hyperparameter specifications. \textit{Mechanistic heterogeneity} underlying what is otherwise ``virtually indistinguishable network activity'' \citep{prinz2004similar} is a hallmark of even relatively ``simple'' biological neural networks. Observing this phenomenon in a small LM suggests similar challenges to the field of mechanistic interpretability at large. Future research could work to identify the \textit{axes of correspondence} for behaviors that are ``conserved'' across model instances \citep{tigges2024llm, trott2025toward, fehlauer2025convergence}, with the goal of linking properties of individual model instances (e.g., size, training data, initial parameters) to the mechanisms and behaviors they exhibit.

\section*{Acknowledgments}
We thank Benjamin K. Bergen, Cameron R. Jones, Catherine Arnett, Tyler A. Chang, James Michaelov, Samuel Taylor, and Yoonwon Jung for valuable feedback, which shaped the direction of this work early on. Pamela D. Rivi\`ere was supported by the UCSD Chancellor's Postdoctoral Fellowship. We additionally wish to thank Stephen J. Hanson for providing generous access to an NVIDIA DGX-H200. We thank the reviewers and editor for helpful suggestions, and for the opportunity to improve our work.

\bibliography{tacl2021}
\bibliographystyle{acl_natbib}

\begin{figure*}[t]
\begin{center}
 \includegraphics[width=0.80\linewidth]{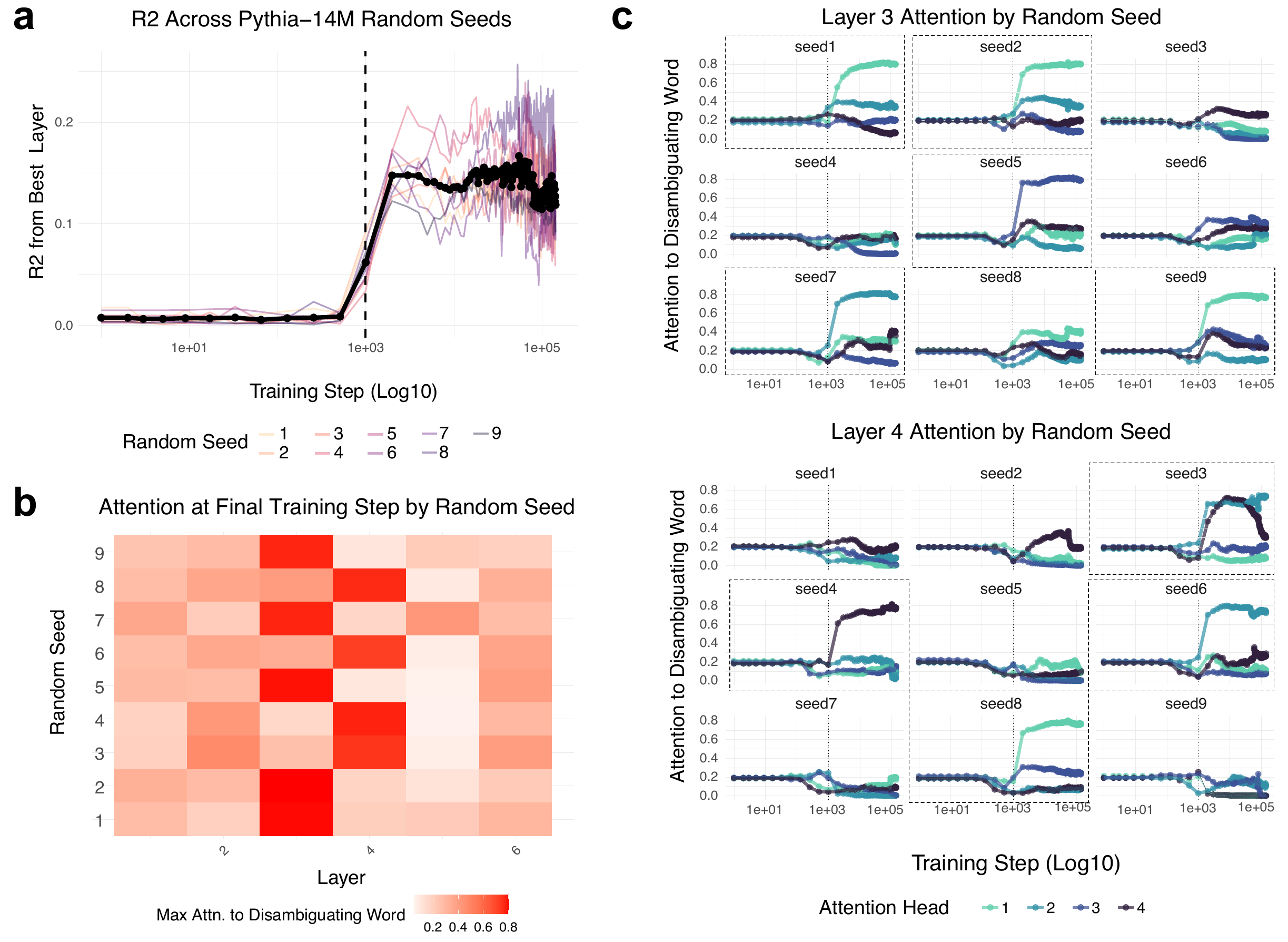} 
\end{center}
\caption{\textbf{Generalizability of Attention Patterns Across Random Seeds} \textbf{(a)} $R^2$ from layer with the maximum $R^2$ by training checkpoint, across all Pythia-$14M$ random seeds. Black curve represents average max $R^2$ across seeds. \textbf{(b)} Maximum attention to disambiguating word in original RAW-C sentences at final training step by random seed (rows) and layer (columns). Warmer colors indicate larger maximum attention scores. The layer that contains critical attention heads varies according to random initialization, but particularly likely to be Layers 3 and 4. \textbf{(c)} \textit{(top)} Layer 3 heads' attention to disambiguating word over the course of pre-training, by random seed. Boxed sub-panels mark seeds containing heads with largest attention to disambiguating word in this layer. \textit{(bottom)} Same as in \textit{top}, but for Layer 4.}\label{fig:generalizability}
\end{figure*}

\section{Appendix}\label{sec:appendix}

\subsection{Generalizability of Attention Patterns Across Pythia-$14M$ Random Seeds}\label{sec:random_seeds}

The analyses above were primarily conducted on two model instances: Pythia-$14M$ and Pythia-$410M$. While the results are encouraging in that both model instances displayed similar developmental trajectories in their disambiguation performance and onset of candidate disambiguation heads, it remains an open question whether these results would generalize \textit{within} a given model architecture but \textit{across random seeds} \citep{zhao2025distributional, trott2025toward, van2025polypythias}. Here, we ask whether the same model (Pythia-$14M$) trained on the same data but with different random random seeds displays regularities across certain \textit{axes of correspondence} \citep{trott2025toward}, e.g., whether candidate attention heads arise at similar \textit{timepoints} or \textit{locations} across models.

\subsubsection{Methods}

We used the nine random seeds released for Pythia-$14M$ \citep{van2025polypythias}. Each model was assessed at all 154 training checkpoints and accessed through the HuggingFace \textit{transformers} library \cite{wolf-etal-2020-transformers}. For tracking changes in disambiguation performance and attention head patterns, we implemented exactly the same procedure used in Phase 1 (see Section \ref{sec:phase1}) for each of the nine random seeds at each training checkpoint. We also introduced a novel behavioral task to assess sensitivity to modifier-noun constructions, which we report on in Appendix Section \ref{sec:appendix-mod-noun}.

\subsubsection{Results}

The developmental trajectories of both disambiguation performance and attention to the disambiguating word were strikingly similar across random seeds. Despite variance in final step performance, each random seed showed sharp changes in $R^2$ at similar checkpoints (\textbf{Figure \ref{fig:generalizability}a}). Similarly, regardless of \textit{where} the ``maximally attentive'' head was located, the largest changes in attention again occurred between steps $1000$ and $2000$ (\textbf{Figure \ref{fig:generalizability}c}). Finally, the random seeds exhibited \textit{bimodality} in where these attention heads developed: in roughly half the seeds, the head with maximal attention to the disambiguating word emerged in Layer 3, while in the other half, it emerged in Layer 4 (\textbf{Figure \ref{fig:generalizability}b}). This is consistent with other work revealing apparent ``bimodality'' across random initializations \citep{zhao2025distributional}.

\subsection{Sensitivity to Modifier-Noun
Constructions}\label{sec:appendix-mod-noun}

We introduced a novel behavioral task to assess changes in each model's knowledge of modifier-noun constructions. For each sentence, we compared the probability assigned by a model to the original modifier-noun construction (i.e., ``wooden beam'') and a swapped version (i.e., ``beam wooden''). We then calculated the log ratio of these probabilities $Log(\frac{p(S_{original})}{p(S_{reversed})})$; a positive log ratio indicated that a higher probability was assigned to the Original ordering, while a negative log ratio indicating a higher probability was assigned to the Reversed ordering. This procedure was repeated across each random seed for Pythia-14M.

\begin{figure}[t]
\center
\includegraphics[width=0.80\linewidth]{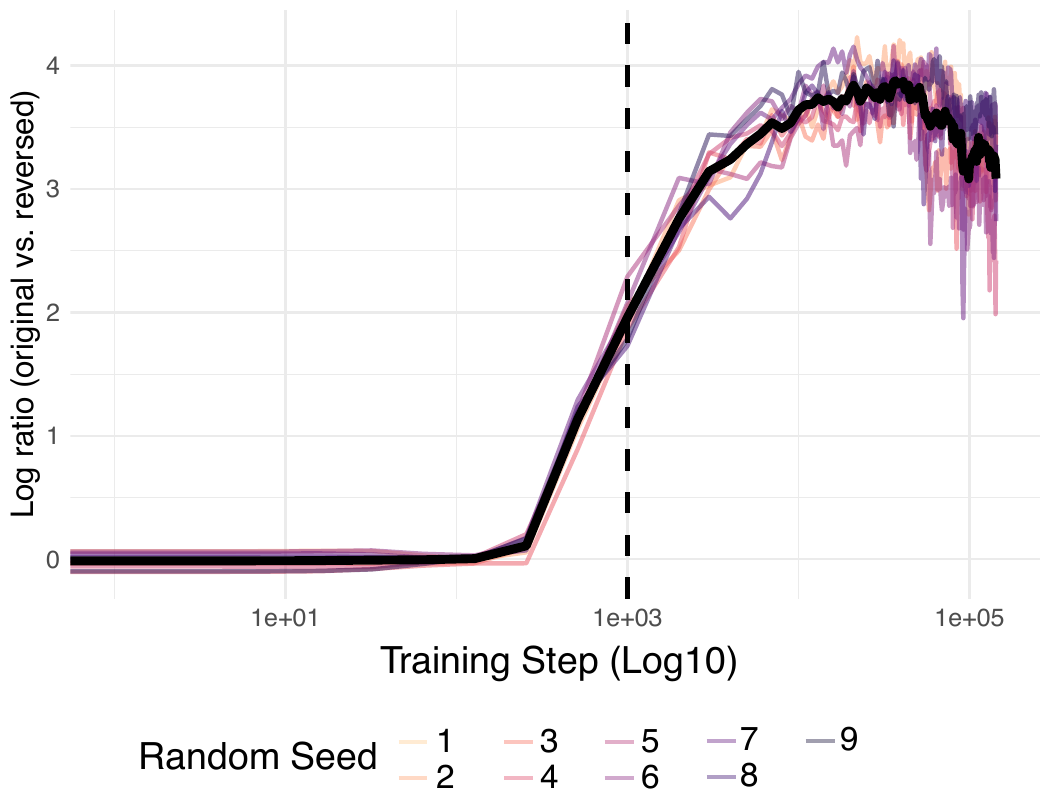} 
\caption{\textbf{Trajectory of sensitivity to modifier-noun constructions is consistent across Pythia-14M random seeds.} Log ratios of the probability assigned to original modifier-noun constructions against that assigned to reversed versions of the constructions. Black curve reflects the average across color-coded random seeds.}\label{fig:appendix-mod-np}
\end{figure}

We observed clear ``phase transitions'' in performance in the modifier-noun task as well, i.e., at steps $512$, $1000$, and $2000$ (\textbf{Figure \ref{fig:appendix-mod-np}}). Moreover, these changes  were highly predictive of changes in disambiguation performance. In a linear mixed effects model with $R^2$ as a dependent variable and Log Ratio and Log Training Step as fixed effects (and Seed as a random intercept), we found that Log Ratio exhibited a positive relationship with $R^2$ $[\beta = 0.04, SE = 0.001, p < .001]$. Moreover, using Akaike Information Criterion (AIC) as a measure of model fit---where a lower AIC value corresponds to a better fit---we found that a regression model with only Log Ratio outperformed a model with only Log Training step ($\Delta AIC > 700$). Together, this suggests that across seeds, there is an inflection point---relatively early in training---involving weight changes to specific attention heads query-key matrices, which results in improved disambiguation performance and enhanced sensitivity to appropriate modifier-noun ordering.

\subsection{Supplementary Figures}
\begin{figure*}[t]
\begin{center}
\includegraphics[width=.85\linewidth]{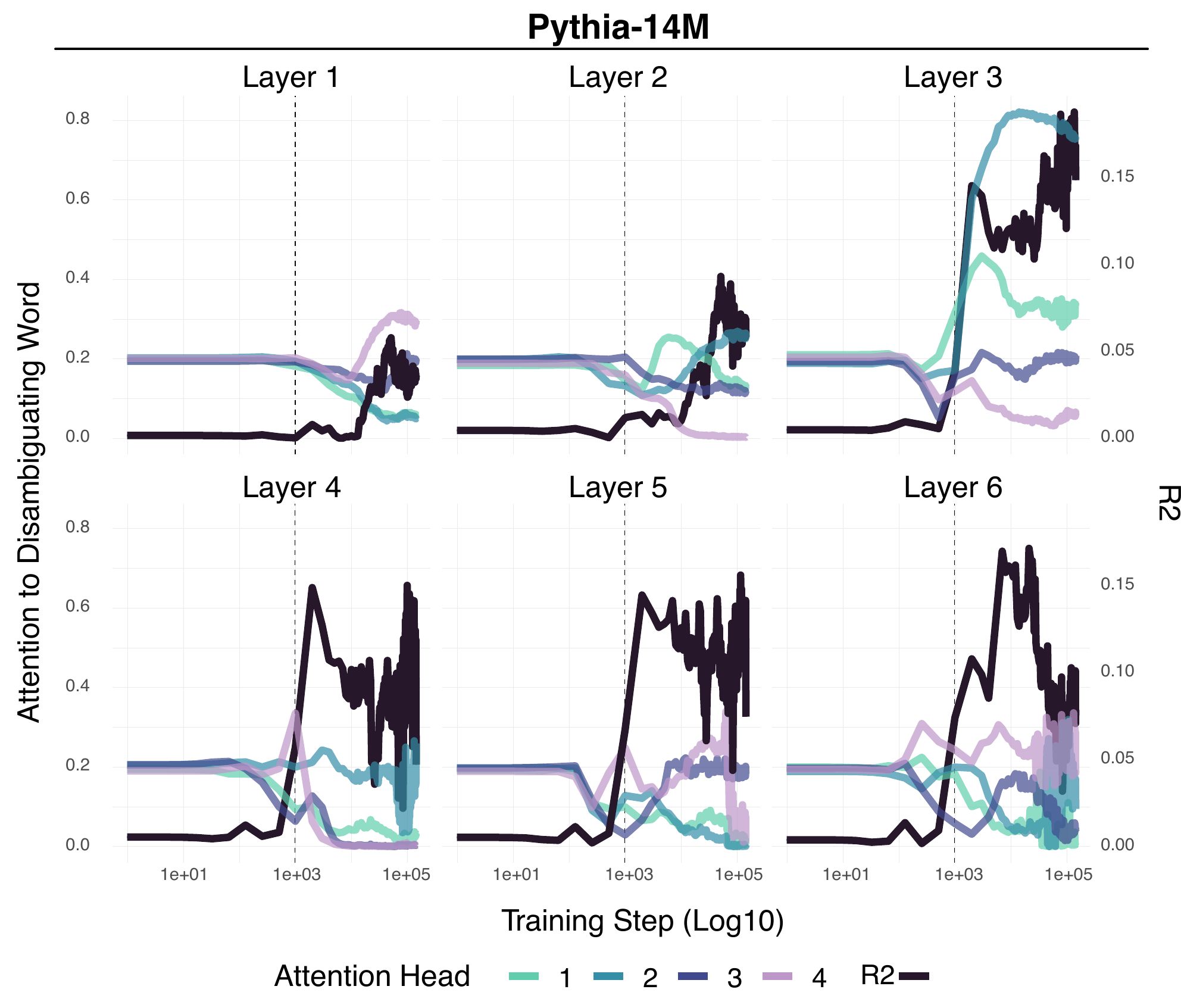}
\end{center}
\caption{\textbf{Companion to Main Text Figure \ref{fig:fig2-identify-candidates}c: Identifying candidate attention heads}. Pythia-$14M$'s attention to the disambiguating word over the course of pre-training, for all color-coded attention heads, for all layers. Superimposed in each sub-panel is the $R2$ (black curve) from each corresponding layer's representations. Layer 3 emerges as the layer containing heads with most pronounced increases in attention time-locked to disambiguation performance.}\label{fig:appendix-fig2ccompanion}
\end{figure*}

\begin{figure*}[t]
\begin{center}
\includegraphics[width=0.9\linewidth]{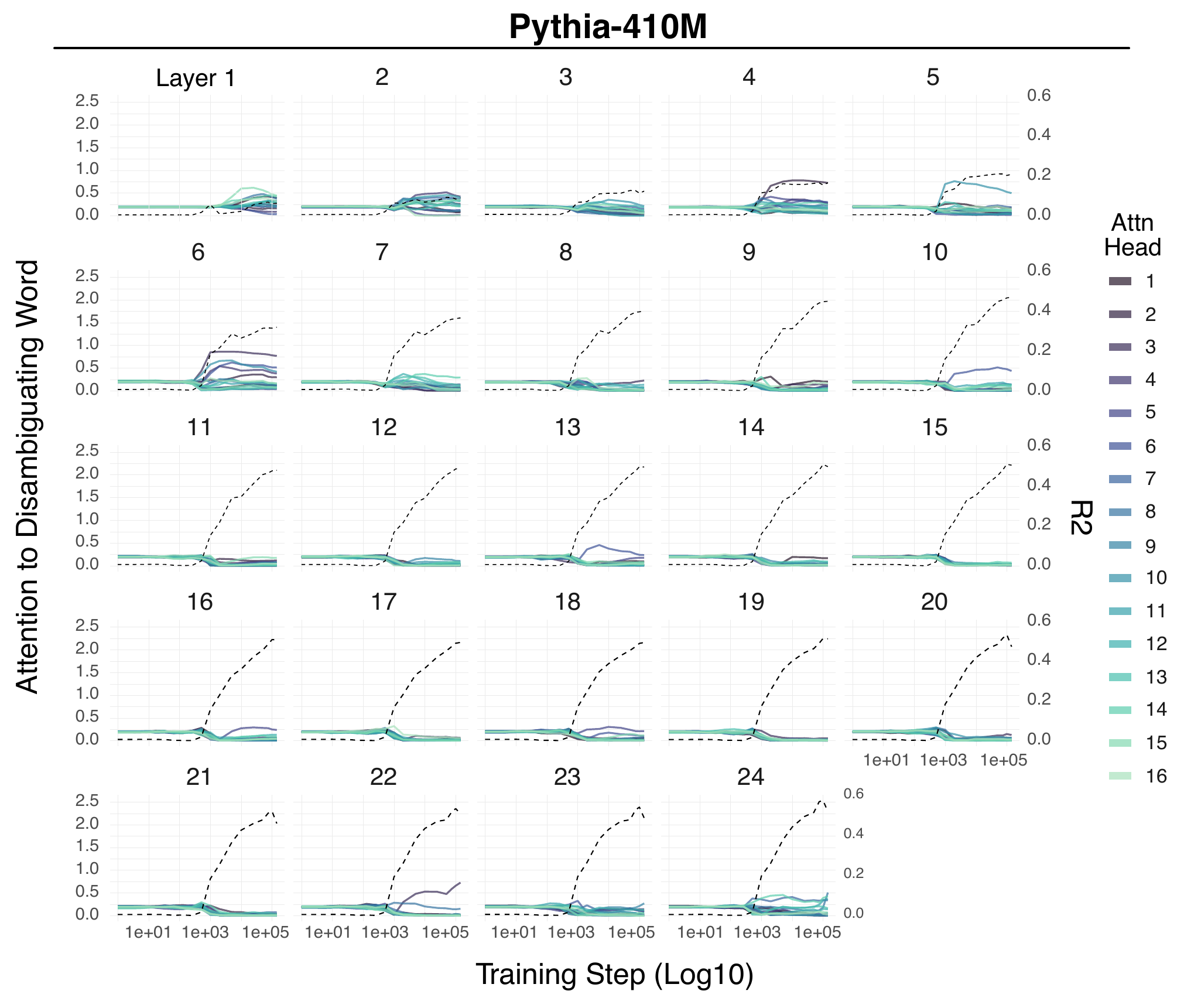}
\end{center}
\caption{\textbf{Companion to Main Text Figure \ref{fig:fig2-identify-candidates}d: Identifying candidate attention heads.} All Pythia-$410M$ layers' attention head trajectories for disambiguating word, over pre-training. Superimposed $R2$ (dashed lines) from each corresponding layer's representations.}\label{fig:appendix-fig2dcompanion}
\end{figure*}

\begin{figure*}[t]
\begin{center}
\includegraphics[width=0.9\linewidth]{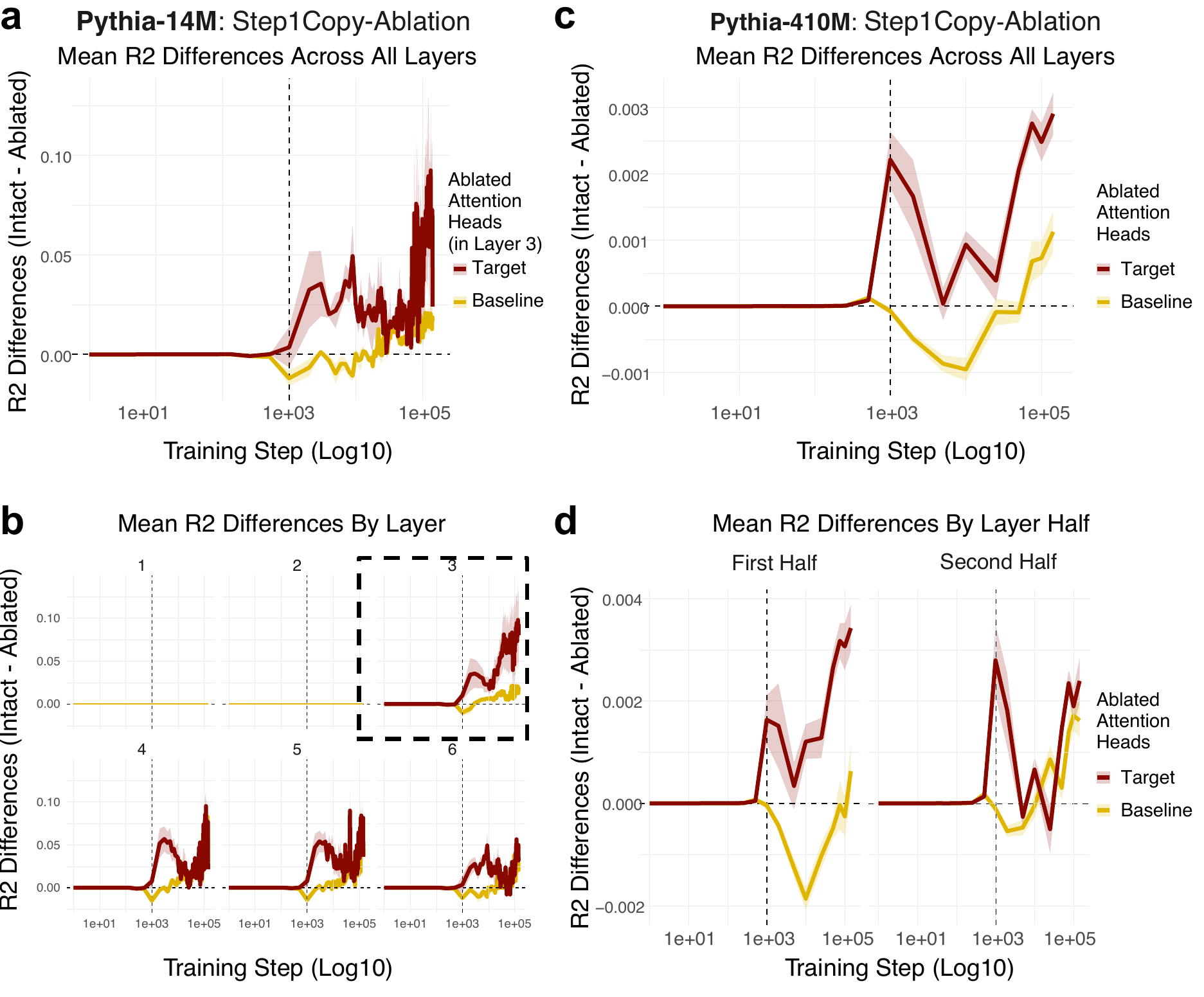}
\end{center}
\caption{\textbf{Companion to Main Text Figure 5: Step-1-Copy Ablations for Pythia-$14M$ and $410M$.} \textbf{(a)} Mean difference in $R2$ across all layers and all combinations of Pythia-$14M$'s target head ablations, for all training steps. Values $>0$ indicate that the intact model's $R2$ exceeded that of the ablated model's $R2$, reflecting causal effect of ablation. Target manipulations refer to ablations of previously-identified Layer 3 heads. Baseline manipulations refer to ablations of Layer 3 heads whose attention to disambiguating words fail to increase with disambiguation performance. \textbf{(b)} Same as in \textbf{a}, but parcelled out by layer, to illustrate localization of ablation effects, which remain robust throughout training in Layer 3. Dashed square marks the only layer (Layer 3) to suffer head ablations. \textbf{(c)} Same as in \textbf{a}, but for Pythia-$410M$. Target head ablations causally decrease model performance. \textbf{(d)} Same as in \textbf{c}, but parcelled out by early versus late layers, to illustrate the selectivity of target-head ablation to earlier layer representations. By the end of training, the effects of target-head ablations do not differ from those of baseline-head ablations.}\label{fig:appendix-fig5companion}
\end{figure*}

\begin{figure*}[t]
\begin{center}
\includegraphics[width=0.9\linewidth]{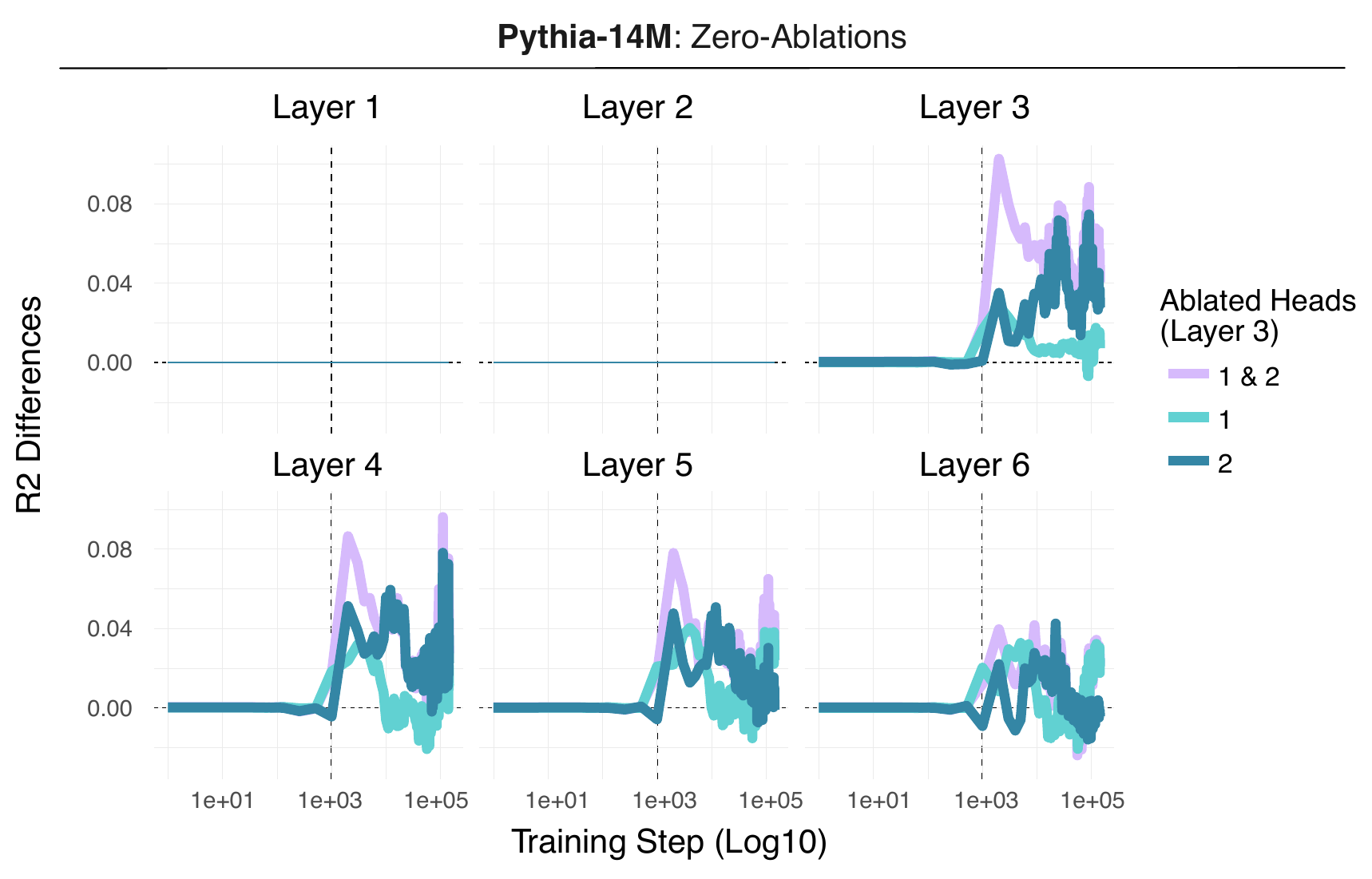}
\end{center}
\caption{\textbf{Companion to Main Text Figure 5: Individual Target Head Zero-Ablation Results, for Pythia-$14M$} \textbf{(a)} Difference in $R2$ (Intact - Ablated) resulting from different color-coded combinations of attention head ablations in Layer 3, across all training checkpoints. Dashed vertical line marks step $1000$. At step $2000$, ablating \textit{both} Heads $(3,1)$ and $(3,2)$ yields much larger performance deficits in Layer 3 representations relative to those of the intact model than ablating each head in isolation. Yet, by the last checkpoints, ablating only Head $(3,2)$ yields nearly the same effect in Layer 3 representations as having ablated both heads. Representations in later model layers, particularly Layer 6, are less affected than those of intermediate layer representations.}\label{fig:appendix-fig5companion-14M-individual-target-heads}
\end{figure*}

We include multiple supplementary figures below, which are meant to serve as companions to select main text figures. Each of these is marked as such, and offers a fuller picture of attention head results for Pythia-$14M$(\textbf{Figure \ref{fig:appendix-fig2ccompanion}}) and $410M$ (\textbf{Figure \ref{fig:appendix-fig2dcompanion}}) relative to $R2$, over the course of pre-training; ablation results for the Step-1-Copy manipulation for both models (\textbf{Figure \ref{fig:appendix-fig5companion}}), which is fully described methodologically in Section \ref{sec:phase3}; and results for zero-ablations of individual target heads and their combination in Pythia-$14M$ (\textbf{Figure \ref{fig:appendix-fig5companion-14M-individual-target-heads}}).

\iftaclpubformat

\onecolumn

\appendix

\fi

\end{document}